\documentclass[10pt,twocolumn,letterpaper]{article}

\usepackage{cvpr}              % To produce the CAMERA-READY version
\definecolor{cvprblue}{rgb}{0.21,0.49,0.74}
\usepackage[pagebackref,breaklinks,colorlinks,allcolors=cvprblue]{hyperref}

\def\paperID{*****} % *** Enter the Paper ID here
\def\confName{CVPR}
\def\confYear{2026}

\title{The 1st Winner for 5th PVUW MeViS-Text Challenge:\\ Strong MLLMs Meet SAM3 for Referring Video Object Segmentation}

\author{
Xusheng He$^{1}$, Canyang Wu$^{1}$, Jinrong Zhang$^{1}$, Weili Guan$^{1,2}$, Jianlong Wu$^{1,2}$, Liqiang Nie$^{1,2}$
\\ 
$^1$Harbin Institute of Technology, Shenzhen, China $^2$Shenzhen Loop Area Institute, China \\ Team: HITsz\_Dragon
}
\begin{document}
\maketitle
\begin{abstract}
This report presents our winning solution to the 5th PVUW MeViS-Text Challenge. The track studies referring video object segmentation under motion-centric language expressions, where the model must jointly understand appearance, temporal behavior, and object interactions. To address this problem, we build a fully training-free pipeline that combines strong multimodal large language models with SAM3. Our method contains three stages. First, Gemini-3.1 Pro decomposes each target event into instance-level grounding targets, selects the frame where the target is most clearly visible, and generates a discriminative description. Second, SAM3-agent produces a precise seed mask on the selected frame, and the official SAM3 tracker propagates the mask through the whole video. Third, a refinement stage uses Qwen3.5-Plus and behavior-level verification to correct ambiguous or semantically inconsistent predictions. Without task-specific fine-tuning, our method ranks first on the PVUW 2026 MeViS-Text test set, achieving a Final score of 0.909064 and a $\mathcal{J}\&\mathcal{F}$ score of 0.7897. The code is available at \href{https://github.com/Moujuruo/MeViSv2_Track_Solution_2026}{MeViSv2\_Track\_Solution\_2026}.
\end{abstract}    
\section{Introduction}
\label{sec:intro}
Video object segmentation (VOS) aims to segment target objects throughout a video while maintaining temporal consistency\cite{pont20172017,ding2023mose,hong2023lvos,xu2018youtube}. In recent years, the task has evolved from semi-supervised settings with a first-frame mask to more general settings, leading to the development of referring video object segmentation (RVOS). Compared with conventional VOS, RVOS requires the model not only to propagate masks over time, but also to understand natural language expressions\cite{ding2023mevis,li2024r,athar2025vicas,li2019expectation, liang2025long}. This makes RVOS a joint problem of visual grounding, temporal understanding, and mask prediction.

The MeViSv2-Text\cite{MeViSv2} track in the PVUW challenge further emphasizes motion-centric referring expressions. Different from appearance-based queries, the target here is often defined by motion, interaction, relative position, or exclusion relations. As a result, the model must reason jointly over spatial appearance and temporal behavior. This setting is particularly challenging in realistic videos, where occlusion, viewpoint changes, cluttered backgrounds, and multiple similar instances are common.

Moreover, unlike static referring expressions, motion-centric queries require comparing observations across time rather than within a single frame. The model must determine not only what the target looks like, but also whether its behavior truly satisfies the description throughout the video.

Recent RVOS methods, including both training-based and training-free paradigms, have shown that a promising strategy is to first use an MLLM to interpret the query and localize the target, and then use a segmentation model to obtain a spatial seed that can be propagated through the video\cite{wu2025survey,zhu2026training,li2025revseg, jiang2026refer}. However, we observe three practical bottlenecks in this line of work. First, many existing systems rely on relatively small open-source MLLMs\cite{bai2025qwen3,lu2025ovis2}, whose video understanding and compositional reasoning remain insufficient for difficult motion expressions. Second, a large portion of prior methods are built upon SAM2\cite{ravi2024sam}, while the newly released SAM3\cite{carion2025sam} provides a stronger segmentation model and a stronger video tracker. Third, many methods still use boxes or special tokens(e.g., $<$SEG$>$)\cite{yuan2025sa2va,bai2024one,tian2025dtos} as the intermediate representation between language understanding and segmentation, which inevitably discards fine-grained referential information in complex scenes.

Motivated by these observations, we build a fully training-free pipeline that couples strong MLLMs with SAM3 for event decomposition, mask generation, and temporally consistent propagation.

\begin{figure*}[t]
\centering
\includegraphics[width=1\textwidth]{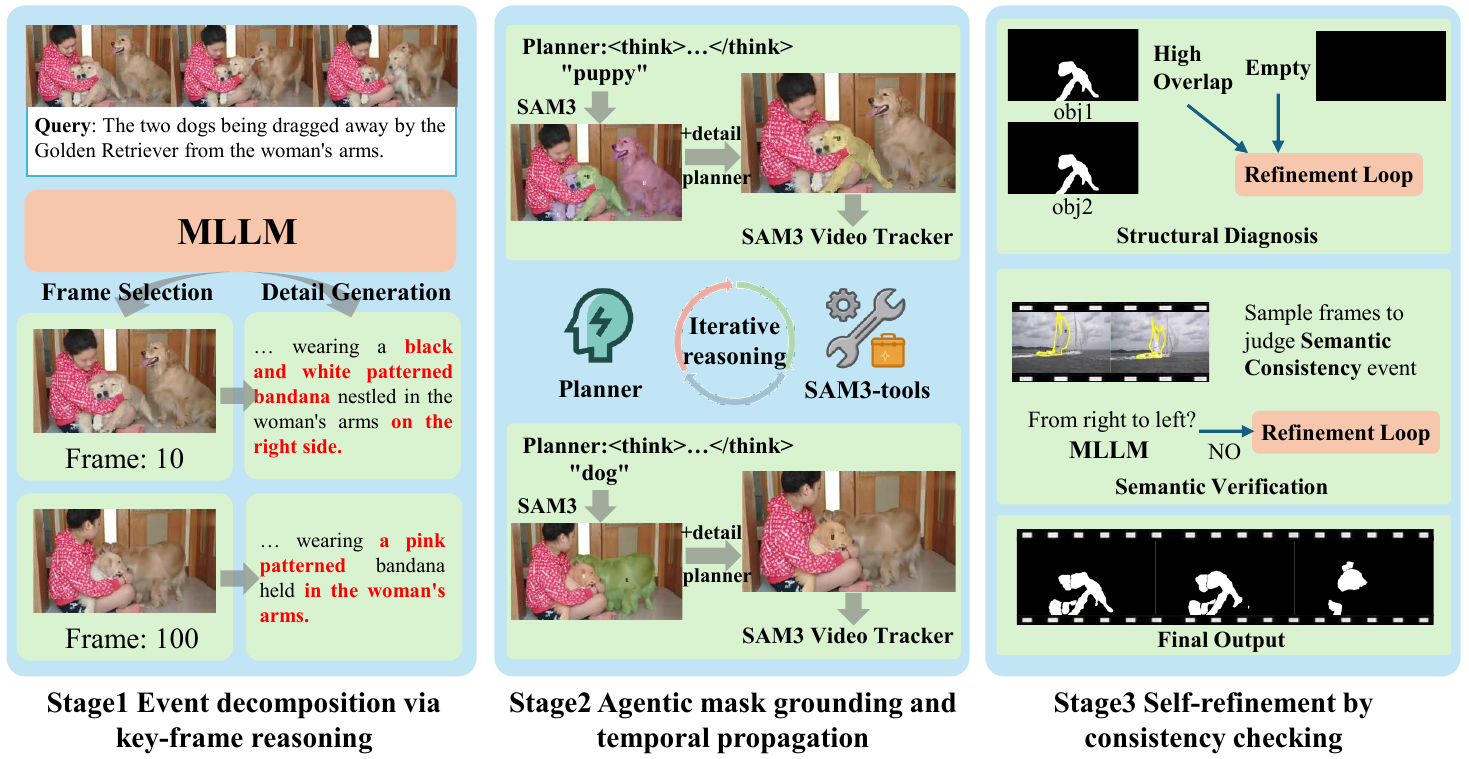}
\caption{Overview of our training-free pipeline for the PVUW MeViS-Text challenge. Stage 1 uses Gemini-3.1 Pro to decompose each target event into instance-level grounding targets, select the clearest frame, and generate a discriminative description. Stage 2 uses SAM3-agent to generate seed masks on the selected frames and propagates them through the video with the SAM3 tracker. Stage 3 further refines ambiguous or semantically inconsistent predictions through mask inspection, description regeneration, and behavior-level verification.}
\label{fig:method}
\end{figure*}

The main designs in this project include:
\begin{itemize}
    \item We develop a stage-wise framework that decomposes motion-centric expressions into instance-level grounding targets before video propagation.
    \item We build a direct mask generation pipeline based on SAM3-agent, which reduces the information loss introduced by intermediate box representations.
    \item We introduce a refinement strategy that improves robustness by revisiting ambiguous or semantically inconsistent predictions.
\end{itemize}

\section{Our solution}
\label{sec:solution}
\subsection{Framework Overview}

Our method follows a three-stage design. We first transform each target event into a set of instance-level grounding targets through MLLM-based event decomposition. We then apply SAM3-agent on the frame where the target object is most clearly visible to obtain precise seed masks. These seed masks are propagated to the full video by the SAM3 tracker. Finally, we introduce a self-refinement stage that revisits ambiguous predictions and keeps the predicted masks semantically consistent with the original event description. By directly generating seed masks rather than intermediate boxes, the proposed pipeline preserves more information for subsequent temporal propagation.

\subsection{Three-Stage Pipeline}

\noindent \textbf{Stage 1: Event decomposition via key-frame reasoning.}
For each video and target event, we use Gemini-3.1 Pro to analyze the video and decompose the event at the instance level. The model first identifies all valid object instances that truly satisfy the event and isolates the central subject from auxiliary referents. It then selects the frame where each target is most clearly visible and generates a discriminative description for each target. This stage converts a difficult video motion expression into a set of image grounding problems and explicitly separates multiple valid instances for subsequent processing in complex scenes.

The key idea of this stage is that the MLLM should not merely summarize the event, but should disentangle it into individual grounding units. Therefore, the generated description emphasizes identifying attributes, local appearance, and scene context that are sufficiently unique in the selected frame. Compared with directly prompting the downstream segmentation model from the raw event description, this decomposition provides a better input for the next stage.

\begin{figure*}[t]
\centering
\includegraphics[width=1\textwidth]{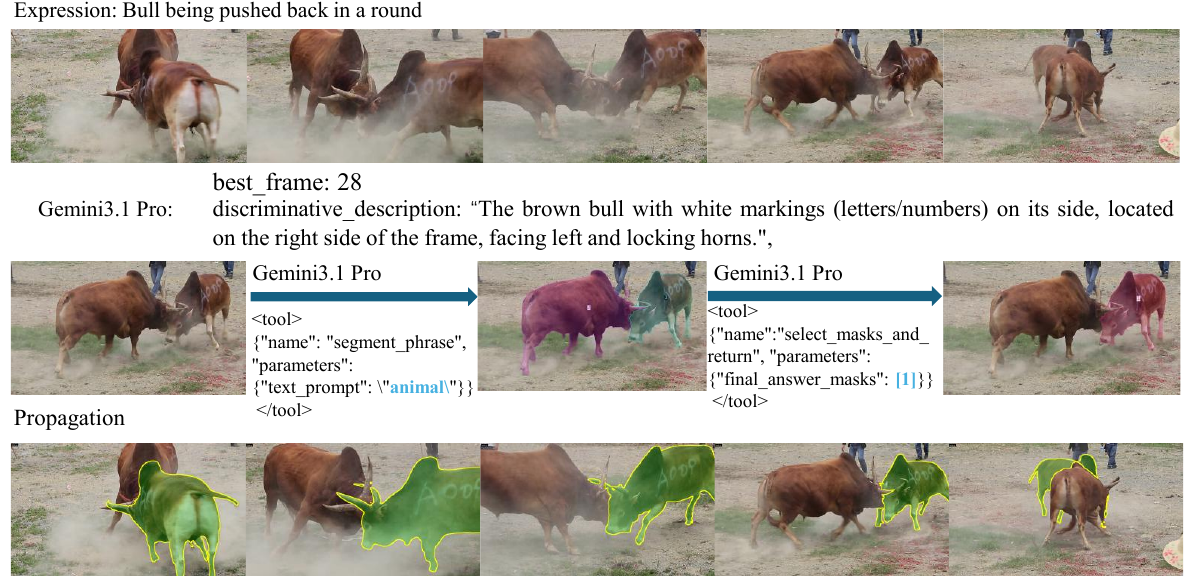}
\caption{Qualitative visualization of our method on the MeViS-Text track. The figure shows representative frames together with the predicted masks, as well as an example of discriminative description generation and SAM3-agent grounding.}
\label{fig:visual1}
\end{figure*}

\noindent \textbf{Stage 2: SAM3-agent grounding and temporal propagation.} After obtaining the key frame and its discriminative description, we use SAM3-agent to generate a precise seed mask on the selected frame. Unlike methods that rely on bounding boxes as intermediate prompts, our method directly predicts a pixel accurate mask. SAM3-agent formulates image segmentation as a reasoning process over multiple rounds. At each round, a multimodal LLM chooses among SAM3 tools according to the current image, the text prompt, and the intermediate results. In our implementation, Gemini-3.1 Pro serves as the planner. Starting from the selected frame and the target description, the planner analyzes the current result, selects the next SAM3 operation, and updates the decision according to the returned masks until a satisfactory target mask is obtained or the target is judged absent.

This method is particularly helpful for complex expressions that require relation understanding, scene reasoning, or disambiguation among similar instances. During mask selection, the planner can continue to use the description throughout the reasoning process. Once the seed mask is obtained, we initialize the official SAM3 video tracker on the selected frame and propagate it in both temporal directions. For expressions with multiple instances, we segment and track each target independently, and merge the masks only after propagation.

\noindent \textbf{Stage 3: Self-refinement by consistency checking.}
Although the above two stages already provide strong results, we observe that the remaining failures mainly come from two sources: insufficient discrimination among multiple similar instances, and semantic inconsistency between the predicted target and the original event. We therefore introduce a refinement stage that automatically detects unreliable cases and reprocesses them.

We begin by inspecting the predicted masks, where empty predictions and highly overlapping masks for distinct targets indicate that the original description is not sufficiently discriminative. We then use Qwen3.5-Plus\cite{qwen35blog} to regenerate a more precise description and rerun the same SAM3-agent grounding procedure. In addition, for expressions with directional or negative constraints, we further perform a behavior-level verification step by sampling frames from the video, highlighting the predicted mask boundaries, and asking an MLLM to judge whether the tracked object is truly consistent with the original event description. If a prediction fails this consistency check, we send it back to the refinement branch for another round of re-grounding. This refinement stage improves the robustness of the overall pipeline.
\section{Experiments}

\subsection{Challenge Description}
The Pixel-level Video Understanding in the Wild (PVUW) Challenge\cite{ding2024pvuw, ding2025pvuw} features three tracks this year. The Complex Video Object Segmentation Track, based on MOSEv2\cite{ding2025mosev2}, focuses on video object segmentation in complex scenarios with challenges such as occlusions, small objects, and crowded environments. The Motion Expression Guided Video Segmentation Tracks, based on MeViSv2-Text and MeViSv2-Audio, explore text- and audio-guided video segmentation using motion expressions. In particular, the MeViSv2-Text track emphasizes the association between motion descriptions and dynamic object behaviors, encouraging the development of more robust pixel-level video understanding methods in realistic videos.

\subsection{Implementation Details}
% \noindent \textbf{Backbone and Pretrained weights}. 
% We utilize the pretrained weight from DinoV2~\cite{dinov2} to initialize the ViT blocks.
% %
% The fusion block is performed every 3 blocks of ViT layes.
% %
% Soft aggregation operation is used to merge the predicted masks in the multi-target scenario.

\noindent\textbf{Inference.} Our method is fully training-free and does not require task-specific optimization on the challenge data. During inference, Gemini-3.1 Pro is used to decompose each target event into instance-level grounding targets and to serve as the planner in SAM3-agent for seed mask generation on the selected frame. Qwen3.5-Plus is used in the refinement stage to regenerate more precise descriptions for problematic cases. After a seed mask is obtained, the official SAM3 tracker propagates it in both temporal directions. For expressions with multiple valid instances, we segment and track each target independently, and merge their masks only after propagation. The refinement stage is triggered by structurally unreliable predictions or semantic inconsistencies, after which the corresponding targets are re-grounded and re-propagated. All submitted results are produced by a single pipeline without additional fine-tuning or model ensembling. All experiments are conducted on a workstation with 2 NVIDIA GeForce RTX 4090 GPUs.

\noindent \textbf{Evaluation Metrics.} 
To comprehensively evaluate the segmentation performance, we use the mean $\mathcal{J}\&\mathcal{F}$ score, which averages the Jaccard $\mathcal{J}$ index for region similarity and the $\mathcal{F}$ score for boundary accuracy. Additionally, we employ No-target accuracy, denoted as N-acc., and Target accuracy, denoted as T-acc., to assess the model's robustness against unmatchable queries and its precision in identifying the correct referents, respectively. Finally, the Final score is computed as the average of the mean $\mathcal{J}\&\mathcal{F}$ score, N-acc., and T-acc. to provide a comprehensive ranking of the overall performance. 
\subsection{Results}

\begin{table}[t]
\centering
\small
\caption{Ranking results (Top 6) in the PVUW 26 MeViS-Text test set. We mark our results in {\color{blue}{blue}}.}
\label{tab:ranking}
\setlength{\tabcolsep}{2mm}{
\renewcommand\arraystretch{1.0}
\begin{tabular}{c | c | c c c c}
\toprule
Rank & Name & $\mathcal{J}\&\mathcal{F}$ & N-acc. & T-acc. & Final \\
\midrule
1 & \textcolor{blue}{HITsz\_Dragon} & \textcolor{blue}{0.7897} & \textcolor{blue}{0.9615} & \textcolor{blue}{0.9759} & \textcolor{blue}{0.909} \\
2 & goodx & 0.7106 & 1.0 & 0.9652 & 0.892 \\
3 & tobedone & 0.713 & 0.9615 & 0.9893 & 0.888 \\
4 & yahooo & 0.7038 & 0.9615 & 0.984 & 0.883 \\
5 & junjie\_zheng & 0.6837 & 0.8846 & 0.9679 & 0.845 \\
6 & rookie7777 & 0.642 & 0.8462 & 0.9679 & 0.819 \\
\bottomrule
\end{tabular}
}
\end{table}

Table~\ref{tab:ranking} reports the ranking results on the PVUW 2026 MeViSv2-Text test set. Figure~\ref{fig:visual1} further shows qualitative examples of our pipeline. Our method ranks first overall with a Final score of 0.909064. In particular, it achieves the best $\mathcal{J}\&\mathcal{F}$ score of 0.7897, outperforming the second-place method by 0.0791. Although our method does not achieve the highest N-acc. or T-acc. individually, it attains the best overall ranking by maintaining a strong balance between target identification and mask quality. These results demonstrate that a carefully designed training-free pipeline, when built upon strong MLLMs and SAM3, can achieve highly competitive performance on motion-centric RVOS.

\section{Conclusion}
\label{sec:Conclusion}

In this report, we presented our solution to the 5th PVUW MeViS-Text Challenge. Our method follows a fully training-free pipeline that combines instance-level event decomposition, SAM3-agent mask grounding, bidirectional SAM3 propagation, and refinement by structural inspection and behavior verification. The core idea is to preserve as much language information as possible before temporal propagation and to directly generate seed masks instead of relying on intermediate box representations. The final results show that this simple but carefully designed pipeline is highly effective, ranking first on the challenge test set. We hope our solution can provide a useful baseline for future research on motion-centric RVOS with strong MLLMs and foundation segmentation models.
{
    \small
    \bibliographystyle{ieeenat_fullname}
    \bibliography{main}
}


\end{document}